\documentclass[12pt]{article}

\usepackage{natbib}
\usepackage[utf8]{inputenc} 
\usepackage[T1]{fontenc}    
\usepackage{hyperref}       
\usepackage{url}            
\usepackage{booktabs}       
\usepackage{amsfonts}       
\usepackage{nicefrac}       
\usepackage{microtype}      
\usepackage[table]{xcolor}         

\usepackage{authblk}

\usepackage{fullpage}

\usepackage{colortbl}
\usepackage{pgf} 
\usepackage{siunitx}

\usepackage{mathtools}
\usepackage{bm}
\usepackage{multirow}

\usepackage{cleveref}

\usepackage[most]{tcolorbox}
\tcbset{
  mytakeaway/.style={
    colback=gray!10,
    colframe=white,
    boxrule=0pt,
    enhanced,
    arc=2pt,
    left=6pt,
    right=6pt,
    top=6pt,
    bottom=6pt,
    fonttitle=\bfseries,
    coltitle=black
  }
}

\newcommand{\lightgbm}{\texttt{LightGBM}}
\newcommand{\GBDT}{\textsc{GBDT}}
\newcommand{\scikitsurvival}{\texttt{Scikit-Survival}}
\newcommand{\Cox}{\textsc{Cox}}

\newcommand{\donor}{\vec{x}_{\mathsf{don}}}
\newcommand{\pat}{\vec{x}_{\mathsf{pat}}}
\newcommand{\waitlist}{\mathcal{W}}

\newcommand{\timeset}{\mathcal{T}}

\newcommand{\donorspace}{\mathcal{D}_{\mathsf{don}}}

\newcommand{\patspace}{\mathcal{D}_{\mathsf{pat}}}

\newcommand{\graftsurv}{\mathsf{GraftSurv}}
\newcommand{\waitsurv}{\mathsf{WaitlistSurv}}
\newcommand{\offeracc}{\mathsf{OfferAcc}}

\newcommand{\deltapot}{\delta_{\mathsf{pot}}}
\newcommand{\vxpot}{\vec{x}_{\mathsf{pot}}}

\newcommand{\policy}{\mathcal{P}}

\renewcommand{\vec}[1]{\bm{#1}}

\title{Policy Optimization for Dynamic Heart Transplant Allocation\thanks{An extended abstract of this paper was presented at the American Heart Association’s (AHA) Scientific Sessions 2025.}}

\author[1]{Ioannis Anagnostides}
\author[2]{Zachary W. Sollie}
\author[2]{Arman Kilic}
\author[1]{Tuomas Sandholm}
\affil[1]{Department of Computer Science, Carnegie Mellon University, Pittsburgh, PA}
\affil[2]{Department of Surgery, Division of Cardiothoracic Surgery, Medical University of South
Carolina, Charleston, SC}

\begin{document}

\maketitle

\begin{abstract}
    Heart transplantation is a viable path for patients suffering from advanced heart failure, but this lifesaving option is severely limited due to donor shortage. Although the current allocation policy was recently revised in 2018, a major concern is that it does not adequately take into account pretransplant and post-transplant mortality. In this paper, we take an important step toward addressing these deficiencies.
    
    To begin with, we use historical data from UNOS to develop a new simulator that enables us to evaluate and compare the performance of different policies. We then leverage our simulator to demonstrate that the \emph{status quo} policy is considerably inferior to the myopic policy that matches incoming donors to the patient who maximizes the number of years gained by the transplant. Moreover, we develop improved policies that account for the dynamic nature of the allocation process through the use of \emph{potentials}---a measure of a patient's utility in future allocations that we introduce. We also show that batching together even a handful of donors---which is a viable option for a certain type of donors---further enhances performance. Our simulator also allows us to evaluate the effect of critical, and often unexplored, factors in the allocation, such as geographic proximity and the tendency to reject offers by the transplant centers.
\end{abstract}

\section{Introduction}

Heart transplantation is the treatment of choice for patients suffering from advanced heart failure---a leading cause of morbidity and mortality worldwide. Unfortunately, the demand for heart transplants far outweighs existing supply~\citep{Cameli22:Donor}. While median post-transplant survival has exceeded $10$ years, this lifesaving option remains severely circumscribed by the number of available donor hearts; indeed, in certain years, the number of waitlisted patients is almost twice the number of patients who proceeded to transplant~\citep{Hsich16:Matching}.

The allocation policy for adult heart transplants in the US was recently revised by the Organ Procurement and Transplantation Network (OPTN), effective October 18, 2018~\citep{Kilic21:Evolving}. In particular, the previous $3$-tier system was further refined into $6$ tiers so as to provide a higher level of granularity based on clinical condition and medical urgency. Retrospective analyses have demonstrated that some of the intended consequences of the policy---such as reducing waitlist mortality---have indeed been materialized~\citep{Liu21:Impact}.

However, many shortcomings still persist. As explained by~\citet{Shore20:Changes}, a major concern is that ``the updated allocation criteria increase the likelihood of transplantation for patients at the highest pretransplant risk yet many of these patients also have the highest post-transplant mortality.'' \citet{Shore20:Changes} recognize that ``a better solution would be to create a heart allocation score,'' which would ``incorporate predicted pretransplant and post-transplant mortality and allow for dynamic changes in listing status based on clinical events.''

\subsection{Our contributions}

In this paper, we respond to this call by developing new policies for heart transplant allocation.

We first develop a new simulator using historical data from the United Network for Organ Sharing (UNOS), which enables us to evaluate the performance of a given policy. To do so, we leverage several machine learning (ML) techniques in order to construct predictive models for the basic components underpinning the simulator---namely, offer acceptance, post-transplant (graft) survival, and 
pretransplant (waitlist) survival; this is the subject of~\Cref{sec:simulator}.

We then rely on our simulator to develop improved policies for performing dynamic heart transplant allocation. We begin by proposing a baseline policy (\Cref{sec:baseline}). It is designed to myopically optimize the number of years gained cumulatively by virtue of the transplants within a certain time frame---this is the main objective we consider in this paper. In our simulations, the myopic policy outperforms the \emph{status quo}---and variants thereof---by a considerable margin (\Cref{sec:statusquo}); the improvement stems from the fact that the myopic policy incorporates (predicted) pretransplant and post-transplant mortality, whereas the existing system fails to adequately take those into account.

Next, in~\Cref{sec:distance}, we examine the impact of two basic factors in the performance of the baseline policy. The first one is the maximum distance allowed between the patient's transplant center and the donor. We find a substantial boost in performance when the maximum distance increases, with diminishing returns occurring at higher distances; this mirrors a finding by~\citet{Papalexopoulos23:Applying} in the context of lung allocation. The second one is more unexplored even beyond heart transplants, and concerns the effect of organ acceptances---centers have the option of rejecting a donor offer if they deem it unacceptable to proceed to transplant. We find that overall efficiency would increase considerably if centers were less prone to rejecting incoming offers. This raises interesting questions with regard to the incentives of individual programs and the role of centralized enforcement going forward.\looseness-1

Moving on, we explore policies beyond the myopic one that account for the dynamic, online nature of the allocation; this is one of the key challenges in this problem. We build on an existing approach in the context of kidney exchange that put forward the use of \emph{potentials}~\citep{Dickerson12:Dynamic,Dickerson15:FutureMatch}; the idea here is that one should take into account the long-term effect of maintaining certain patients in the pool, quantified by a metric referred to as the ``potential.'' The parameters of the potential are not known beforehand, so we use as training existing trajectories to estimate those parameters based on simulations. As a proof of concept, we adapt the myopic policy by incorporating potentials parameterized by patients' blood types---a patient with blood type AB is easier to match in the future, so one should, in certain cases, prioritize (compatible) patients with blood type O. We report significant performance gains compared to the baseline policy by incorporating potentials (\Cref{sec:potentials}). This approach has the additional benefit of being particularly interpretable.

Finally, we explore policies that have the ability to allocate multiple donors together (\Cref{sec:batch}). Existing approaches are predicated on dispatching incoming donors immediately. Yet, certain types of donors---namely, ones that did not succumb to circulatory death; non-DCD donors---can be maintained in approximately stable state for a small time window, which we conservatively take it to be $1$ to $2$ days. By allocating multiple donors at once, we develop myopic policies that instead rely on maximum matching to identify better donor-patient pairs; the idea here is that one can partially alleviate the online nature of the problem. We again report considerable improvements compared to the baseline policy even for modest batch sizes.

\subsection{Related research}

Our approach in~\Cref{sec:potentials} builds on prior literature in the context of dynamic \emph{kidney exchange}. In particular, while fielded exchanges are typically myopic---in that only the current pool of pairs is considered during planning---\citet{Dickerson12:Dynamic} introduced the idea of using \emph{potentials} to account for the dynamic nature of the allocation. This approach was further pursued by~\citet{Dickerson15:FutureMatch}, who introduced the framework of~\textsc{FutureMatch}. In kidney exchange simulations, \textsc{FutureMatch} was able to outperform the myopic policy by considering the long-term composition of the pool. We adapt their approach for heart allocation in~\Cref{sec:potentials}, which nevertheless has many distinct features compared to kidney exchange. (In kidney exchange, willing but incompatible donor-patient pairs can exchange organs through barter cycles or chains triggered by altruistic donors; a key difference with heart transplant allocation is that donating one kidney is a viable option, so in kidney exchange donors are part of the active pool.)

\citet{Papalexopoulos23:Applying} recently developed the continuous distribution framework for lung transplant allocation. This is currently the nationwide policy in place for that organ, and is expected to be extended to heart allocation in the future. In that framework, a composite allocation score (CAS) is determined by taking into account several factors, such as geographical proximity (associated with placement efficiency), medical urgency, and whether the patient is pediatric. In particular, CAS is defined as a linear function of those variables, so the problem reduces to estimating those coefficients; this has the advantage of being interpretable and low dimensional, thereby making optimization tractable even though the underlying optimization landscape could be challenging. \citet{Papalexopoulos23:Applying} then treat lung transplant allocation as a multi-objective problem, where one needs to balance between different goals such as placement efficiency, fairness, and patient survival.

\citet{Berrevoets21:Learning} introduced a framework for organ allocation whereby organs are grouped into types and incoming patients are assigned to different priority queues based on the type. This approach also intends to account for the dynamic nature of the problem and improve upon myopic allocations. Essential for such approaches is the development of models for learning to predict donor-recipient compatibility. A notable recent contribution here is the paper of~\citet{Xu21:Learning}, which addresses the problem of estimating transplant outcomes under counterfactual matches not observed in the data; this is also a key challenge in our problem. Relatedly, \citet{Yoon17:Personalized} develop a framework for making prognosis on a broad pool of potential recipient-donor pairs; it divides the feature space into clusters and constructs different predictive models for each cluster. \citet{Qin21:Closing} introduced a framework to understand the factors that drive decisions on organ offers and applied it to liver transplantation data from OPTN. They employ this framework to determine which criteria are most relevant to clinicians for organ offer acceptance, and understand practices from different centers.

Returning to the existing policy for heart allocation, it is worth noting that another widespread issue relates to the manipulability of the existing system. \citet{Shore20:Changes} elaborate on this point: ``Like the prior allocation scheme, the new criteria still continue to define highest urgency based, in part, on treatments administered rather than by disease severity. These may not always be aligned. Patients in cardiogenic shock can be treated with inotropes or by using temporary mechanical circulatory support devices (MCSDs) or durable LVADs at the treating physician’s discretion but with different implications for listing status. Because use of temporary MCSDs would result in a higher listing status, we anticipate greater utilization of temporary MCSDs including extracorporeal membrane oxygenation.'' As a result, this incentivizes patients to engage with multiple transplant centers in search of a more favorable listing status, which goes against the mandate to equitable access. As a result, developing objective criteria for ascertaining a patient's medical urgency is of great value; our approach in this paper can be seen as a step in that direction.
\section{The simulator and our predictors}
\label{sec:simulator}

The basis for performing policy optimization is a reliable \emph{simulator} for heart transplant allocation. We develop such a simulator in this paper, which is the subject of this section. In particular, in \Cref{sec:offeracceptance,sec:graftsurvival,sec:waitlistsurvival}, we use several ML techniques to construct new prediction models for some crucial components of the simulator. The training of our models is based on real-world, historical data made available by UNOS, going back to $1987$.

Throughout this paper, we consider exclusively the problem of heart transplant allocation, although most of the techniques we develop---both on the ML side in enhancing predictive performance and the optimization side---are more general and can be adapted to other organs as well; in particular, we do not consider thoracic patients who are simultaneously listed for lung transplant.

Furthermore, we consider only adult donors and patients---that is, individuals aged at least $18$ years old; for patients, this age criterion is applied at the time they first get listed for a heart transplant.

An overview of the main prediction models comprising the simulator is given in~\Cref{table:prediction}.

\begin{table}[ht]
\centering
\footnotesize
\caption{Overview of our prediction models.}
\begin{tabular}{lcccl}
\toprule
& \textbf{Model} & \# \textbf{Examples} & \# \textbf{Covariates} & \textbf{Performance} \\
\midrule
Offer acceptance (\Cref{sec:offeracceptance})  & \GBDT & $756,299$ & $100$ & AUROC: $0.895 \pm 0.001$ \\
Graft survival (\Cref{sec:graftsurvival})  & \Cox & $60,055$ & $120$ & C-index: $0.600 \pm 0.004$ \\
Waitlist survival (\Cref{sec:waitlistsurvival}) & \Cox & $120,282$ & $30$ & C-index: $0.726 \pm 0.005$ \\
\bottomrule
\end{tabular}
\label{table:prediction}
\end{table}

\subsection{Offer acceptance}
\label{sec:offeracceptance}

A policy is triggered by the arrival of a new (deceased) donor, whereupon it proceeds by identifying a candidate from the waitlist according to some selection criterion. This decision is a function of the waitlist, the policy's internal state---depending on the history of observations so far---and the characteristics of the donor. Now, in practice, once a donor-patient pair has been selected, it does not always proceed to transplant. Instead, this produces an \emph{offer} to the transplant center of the patient, which may or may not accept that offer; in fact, it turns out that most offers end up being declined, which is why the number of offers far outweighs the number of pairs proceeding to transplant (\Cref{table:prediction}). (If the offer gets rejected, the policy proceeds recursively by considering the rest of the patients in the waitlist; for this reason, the policy should determine an entire priority ordering over the waitlisted patients.)

The first component of the simulator deals with this exact question: will a given donor be deemed acceptable to proceed to transplant by the center of a given candidate? We refer to this as the \emph{offer acceptance} problem. It is a binary classification task: given the features of a donor-patient pair, output a number in $[0, 1]$, to be interpreted as the probability of acceptance.

Using historical data from UNOS, we constructed $756,299$ examples for this problem, each labeled `$0$' or `$1$'---indicating whether that pair ended up proceeding to transplant; to produce this dataset, we followed the data cleaning process described by the Scientific Registry of Transplant Recipients (SRTR)~\citep{SRTR:Offer}. Each example comprises $100$ covariates (excluding the label), capturing characteristics of both the donor and the candidate. We have included most of the ones used by SRTR, together with additional ones so as to enhance the discriminatory power of the model.

\paragraph{The model} To construct a prediction model, we make use of~\lightgbm~\citep{Ke17:LightGBM}. It is an efficient implementation of \emph{gradient boosting decision trees (\GBDT)}, which is a state-of-the-art ensemble model of decision trees~\citep{Friedman01:Greedy}. In particular, we trained a regression model using $200$ estimators and learning rate $0.1$. The dataset was randomly split into $70\%$ training, while the rest was in turn equally partitioned into validation and testing set; we used the validation set during the training phase to regulate overfitting. We partitioned the covariates to categorical and numerical. The former were imputed according to the most frequent, while the latter based on the mean. The numerical values also underwent standard rescaling. We handled the categorical ones via the internal implementation of~\lightgbm. It is worth noting that the SRTR model manually encodes the categorical data, but in so doing it loses a lot of information because most of the classes get lost. We refrain from doing this to extract as much information as possible from the data.

\paragraph{Evaluation} Repeating this process $5$ times, we report a (macro-average) AUROC---the area under the receiver operating characteristic curve~\citep{Fawcett06:Introduction}---with mean $0.895$ and (population) standard deviation of $0.001$ (\Cref{table:prediction}). This metric shows that this model enjoys high fidelity, and is on par with a recent model that leverages deep neural networks with embedding layers~\citep{Anagnostides25:Machine}.


\subsection{Post-transplant outcomes: graft survival}
\label{sec:graftsurvival}

Our next predictor concerns post-transplant outcomes, which is essential for improving heart transplant allocation outcomes. In particular, our goal is to predict \emph{graft survival}---duration of a functioning transplant---of a given patient coupled with an incoming donor.

A key challenge in this problem is \emph{right censorship}; namely, for many heart transplants we do not observe the actual duration. This can happen either because the patient has still a functioning transplant at the time of the last observation or because that patient stopped reporting back to the center; when the event---in this case, graft failure---has not yet been observed, we say that there is censorship. This renders many usual ML approaches of little use.

\paragraph{The model}

To address this, we employ \emph{Cox regression ($\Cox$)}, a model that accounts for right censorship. It hinges on the so-called \emph{proportional hazards assumption}~\citep{Cox72:Regression}, which postulates that the \emph{hazard function} can be expressed as $h(t | \vec{x} ) \coloneqq h_0(t) \exp ( \vec{\theta}^\top \vec{x} )$; here, $\vec{x} \in \mathbb{R}^d$ is vector of covariates; $h_0(\cdot)$ the \emph{baseline} hazard; and $\vec{\theta} \in \mathbb{R}^d$ is the underlying parameter vector. If $S(t|\vec{x})$ is the \emph{survival function}---the probability of surviving at least up to time $t$---we have
\begin{equation}
    \label{eq:surv-func}
  S(t | \vec{x}) = \exp \left( - \exp ( \vec{\theta}^\top \vec{x} ) \int_{\tau = 0}^t h_0(\tau) d \tau \right).  
\end{equation}
We set up a Cox regression model using \scikitsurvival, a Python module for performing survival analysis~\citep{Polster20:Scikit}. We constructed $60,055$ examples, each comprising $120$ covariates---corresponding to both donor and patient characteristics; we again significantly expand on the covariates taken into account in the SRTR model~\citep{SRTR:Posttransplant}. The baseline hazard function is computed using Breslow's estimator~\citep{Breslow75:Analysis}. We also use a regularization parameter $\alpha \coloneqq 0.1$ for the ridge regression penalty. The imputation and scaling process is as in~\Cref{sec:offeracceptance}.

\paragraph{Evaluation} As is standard in survival analysis, we evaluate the performance of our model via the \emph{concordance index (C-index)}~\citep{Harrell96:Multivariable}, which is the proportion of all \emph{comparable} pairs in which the predictions and the actual outcomes agree; in the presence of right censorship, two examples are ``comparable'' if i) both experienced the event at different times, or ii) the one with the shorter observed survival time experienced the event---implying that the other subject outlived the former. (A pair is not comparable if the events were experienced at the same time.) In this context, a $C$-index of $0.5$ corresponds to random guessing; the closer it is to $1$, the more reliable the model is.

Repeating the training process $5$ times, we report a mean $C$-index of $0.600$ and standard deviation $0.004$. This performance is on par with the state-of-the-art models for this problem (\emph{e.g.},~\citealp{Lee18:Deephit,Nilsson15:International,Aleksova20:Risk}). Predicting graft survival following a heart transplant is a challenging task, which is why the best available models only attain a relatively modest performance. It is worth mentioning that there is ample work on predicting $1$- or $3$-year outcomes, with existing models performing reasonably well~\citep{Ayers21:Using}, but this does not suffice for our purposes since we need an estimate for the entire graft survival. 

Using Cox regression, we can easily extract the \emph{median} predicted graft survival of a given donor-patient pair using~\eqref{eq:surv-func}. We estimate based on the median and not the expectation in accordance with common practices in this area; among other reasons, the expected value requires full knowledge of the tail of the distribution, which is hard to obtain.

\subsection{Pretransplant outcomes: waitlist mortality}
\label{sec:waitlistsurvival}

Optimizing for post-transplant outcomes alone is not enough. For example, a relatively healthy patient may be expected to have a prolonged graft survival, but that is clearly not the right match if that patient is expected to survive long even without the transplant. This is where pretransplant outcomes---that is, waitlist mortality---comes into play. It allows us to account for the medical urgency of each patient in the waitlist so as to allocate the organ optimally.\footnote{A recent work by~\citet{Zhang24:Development} develops a new risk score for identifying medically urgent candidates at the highest risk of death without transplantation, improving over the 6-status system currently in place (\emph{cf.}~\Cref{sec:statusquo}).} Predicting waitlist mortality is a similar problem to graft survival (\Cref{sec:graftsurvival}). Right censorship is especially commonplace; in particular, we treat patients that eventually underwent transplant as censored at the time of the transplant. Our model is based on the state of the patient at the time of listing. We follow a similar training process as in~\Cref{sec:graftsurvival}. The total number of examples is $120,282$, each comprising $30$ covariates. Repeating the training process $5$ times, we obtain a Cox regression model with a mean $C$-index of $0.726$ and standard deviation $0.005$.

\subsection{Putting it all together}

Armed with these basic modules, we simulate a policy by generating real-world trajectories based on the UNOS historical data. The simulation is carried out for a specified, relatively short, time frame. Throughout the simulation, patients can i) enter the waitlist, ii) withdraw from the waitlist either because they proceeded to transplant or passed away, and iii) update their status because of the progression of their condition. Dynamic covariates that affect waitlist mortality and get dynamically updated include bilirubin, cardiac output, serum creatinine, and several blood pressure measures. In cases where the evaluated policy deviates from the \emph{status quo}, it is possible that we no longer have information on the progression of some patients. We make the simplifying assumption that such patients remain stable for the duration of the simulation. Developing models for simulating patient progression is an important but challenging direction for future work.

\section{Policy optimization for heart transplant allocation}

We are now ready to leverage our simulator to develop improved policies for heart allocation. We begin by introducing a simple myopic policy in~\Cref{sec:baseline}. It is used as the basis for evaluating the significance of certain critical factors in the allocation (\Cref{sec:distance}), and is next compared with the \emph{status quo} policy (\Cref{sec:statusquo}). \Cref{sec:potentials} develops a non-myopic approach that accounts for the dynamic nature of the allocation using potentials. Finally, \Cref{sec:batch} explores the potential benefits in allocating donors in small batches.\footnote{All experiments were run on a 64-core AMD EPYC 7282 processor. Each run was allocated one thread with a maximum of 48GBs of RAM.}


\subsection{A baseline myopic policy}
\label{sec:baseline}

Let $\waitlist^{(t)}$ be the waitlist at some time $t$---consisting of a set of active patients---and an incoming donor $\donor \in \donorspace$. For a patient $\pat \in \patspace$, we define $\graftsurv : \donor \times \pat \to \mathbb{R}_{\geq 0}$ and $\waitsurv : \pat \to \mathbb{R}_{\geq 0}$ in accordance with~\Cref{sec:graftsurvival,sec:waitlistsurvival}. Our goal is to maximize
\begin{equation}
    \label{eq:objective}
    \sum_{t \in \timeset} \left( \graftsurv(\donor^{(t)}, \pat^{(t)} ) - \waitsurv( \pat^{(t)} ) \right),
\end{equation}
where $\timeset$ is the set of time indices corresponding to the arrival of donors; $\donor^{(t)}$ is the incoming donor at time $t \in \timeset$; and $\pat^{(t)} \coloneqq \policy ( \waitlist^{(t)}, \donor^{(t)})$ is the patient selected by the policy $\policy$ to proceed to transplant. The objective~\eqref{eq:objective} captures the cumulative number of (estimated) years added by virtue of the transplants. We consider~\eqref{eq:objective} to be the most natural objective for this problem. That is not to say it is the only natural one; for example, accounting for fairness considerations between different demographic groups is an important direction.

In this context, perhaps the most natural policy that takes into account both the graft survival and waitlist survival is the one that myopically selects the patient that maximizes the difference
\begin{equation}
    \label{eq:delta}
   \delta^{(t)} : \patspace \ni \pat \mapsto \graftsurv(\donor^{(t)}, \pat ) - \waitsurv( \pat ), 
\end{equation}
where $\pat \in \waitlist^{(t)}$; \eqref{eq:delta} is the (estimated) number of years by virtue of the transplant.\footnote{One could adapt~\eqref{eq:delta} to further prioritize critically ill patients by multiplying the waitlist survival by a factor $W \geq 1$.} In particular, a selection is made if only if $\delta^{(t)} > 0$ for at least one patient. We shall refer to this as the \emph{myopic policy}. As we shall see in~\Cref{sec:statusquo}, this simple policy already outperforms the \emph{status quo} by a considerable margin. 

Naturally, only patients who are compatible with the incoming donor in terms of blood type are to be considered. A more lenient constraint concerns the distance between the patient's transplant center and the donor, which is discussed in more detail next.

It is worth mentioning that this myopic policy, leveraging Cox models for estimating graft and waitlist survival, is akin to what is employed for liver allocation in the UK~\citep{Allen2024:Transplant}. In that context, the difference~\eqref{eq:delta} is referred to as ``transplant benefit score.''

\subsection{The effect of offer rejections and geographic proximity}
\label{sec:distance}

Having laid down a basic baseline policy, we proceed by first examining two central questions concerning the allocation: how does i) offer acceptance and ii) maximum distance affect the overall performance of the myopic policy?

\begin{figure}[!ht]
    \centering
    \includegraphics[scale=0.5]{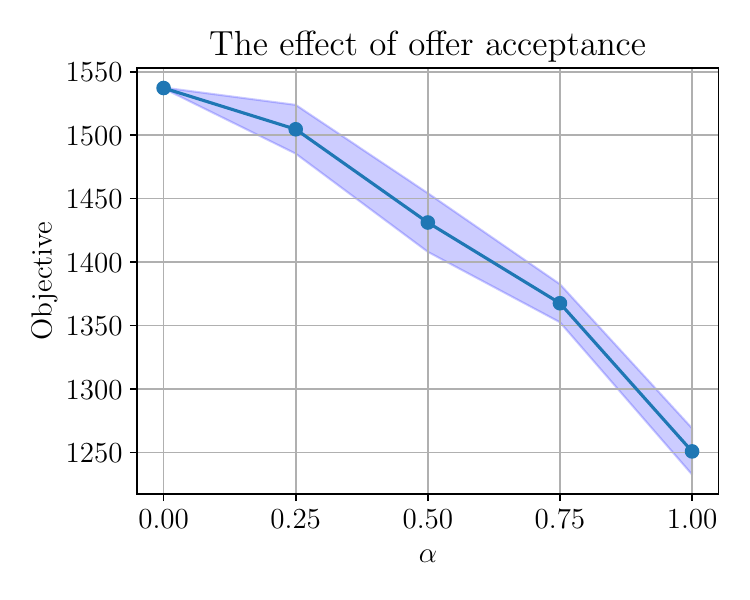}
    \includegraphics[scale=0.5]{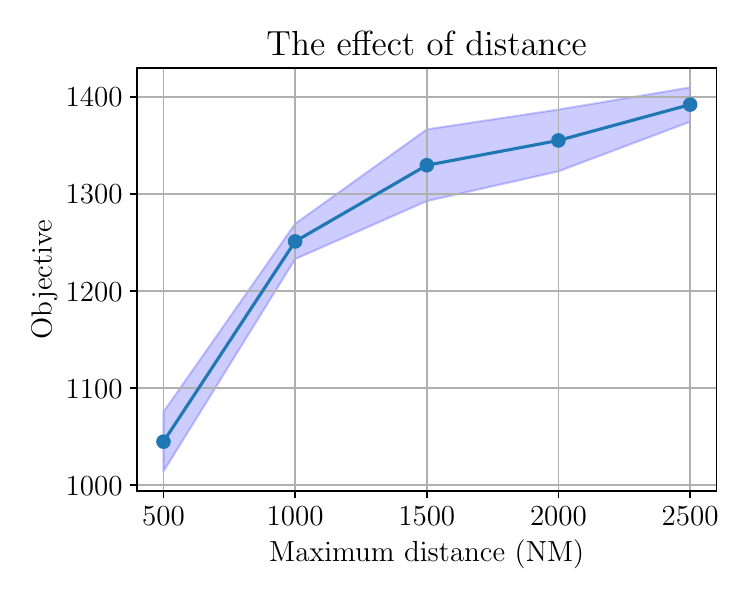}
    \caption{The effect of offer acceptance (left) and maximum distance (right) in the performance of the myopic policy. If $p \coloneqq \offeracc(\donor, \pat)$ is the probability of offer acceptance, the figure on the left shows  what would happen if $p$ is replaced by $p^\alpha$. The error bars indicate the standard deviation obtained from $10$ independent runs.}
    \label{fig:alpha-distance}
\end{figure}

\paragraph{Offer acceptance} Offer acceptance was discussed earlier in~\Cref{sec:offeracceptance}. We saw that individual transplant centers have the option of rejecting an incoming donor. The decision making behind those choices is multifaceted and not well understood. One interesting question is about the effect of eliminating that choice from the transplant centers. That is, one can imagine an alternative system in which decisions are forced to the centers. \Cref{fig:alpha-distance} (left) shows that this would have a drastic improvement in overall performance based on our simulations. That is, we find that offer acceptance is often misaligned with post-transplant and pretransplant outcomes.

To be more precise, this experiment examines the effect of replacing $p \coloneqq \offeracc(\donor, \pat)$ by $p^{\alpha}$ for some parameter $\alpha \in [0, 1]$. By convention, we take $p^{\alpha} = 1$ when $\alpha = 0$, which means that the donor offer is always accepted. The existing system, wherein $\alpha = 1$, lies at the other end of the spectrum. We find an improvement of roughly $23 \%$ when going from $\alpha = 1$ to $\alpha = 0$.

\begin{tcolorbox}[mytakeaway]
\textbf{Takeaway:} Making the transplant centers more permissive in accepting donor offers could have a drastic improvement in overall performance based on our simulations.
\end{tcolorbox}

To our knowledge, there has been no prior research examining why centers are prone to rejecting offers. One plausible hypothesis is that the high rejection rate is to a large extent caused by misaligned incentives between the individual transplant centers and efficiency at the national level. Transplant centers are evaluated twice a year based on attained outcomes as a function of estimated risk.\footnote{More information is given by the~\citet{srtr_psr}.} The current evaluation system may prioritize quality over quantity although it tries to correct for the difficulty of the case to an extent. Also, the current approach potentially creates an incentive for centers with strong recent performance to inefficiently accept riskier cases and, conversely, inefficiently focus on low-risk cases only if they have been underperforming. Investigating these issues in more depth is an interesting direction for future research.

\paragraph{The effect of distance} In what follows, we fix $\alpha = 1$---per the existing policy. The second crucial parameter we examine is the maximum distance allowed between a donor and a patient's transplant center. Geographic proximity is an important part of the equation in organ allocation. Most important, prolonged ischemic time---the duration the donor heart remains without blood supply---increases graft failure risk, which is reflected in our graft survival model. A secondary reason has to do with the logistical overhead involved with longer distances. The effect of the maximum distance (in nautical miles) is shown in~\Cref{fig:alpha-distance} (right). Based on our simulations, we report a massive improvement gain when going from $500$ to $1000$ nautical miles (roughly $20 \%$). Substantial is also the effect of going from $1000$ to $2000$ nautical miles (roughly $8 \%$), although the benefit is more modest in that range.

\begin{tcolorbox}[mytakeaway]
\textbf{Takeaway:} It is essential to consider pairs within radius at least $1000$ nautical miles; increasing the distance further gradually leads to diminishing returns based on our simulations.
\end{tcolorbox}

We caution that this takeaway is predicated on the reliability of the graft survival model. Most examples in the dataset correspond to pairs within relatively close proximity---this is a byproduct of the way the \emph{status quo} policy makes decisions, prioritizing based on the distance. This means that the survival analysis is potentially less reliable when higher distances are considered. While there is some prior work that addresses such biases (for example, we point to~\citealt{Curth21:SurvITE}), it is not clear whether such approaches are effective in handling the specific selection bias present in heart transplant allocation.

\subsection{Comparison of the myopic policy versus the \emph{status quo}}
\label{sec:statusquo}

Moving forward, we now compare the myopic policy with the one currently deployed in the US---which will be referred to as the \emph{status quo}. The latter policy is a tier-based system. Patients are placed into tiers based on i) geographic proximity from the donor, ii) blood type match, and iii) the \emph{status} of the patient, which is a measure of the severity of their condition. As we alluded to in the introduction, as of Octover $2018$, a patient is assigned a status from $1$ to $6$, with $1$ being the most severe. Patients that belong to the same tier are sorted based on their waitlist time, with priority given to those who were listed earlier; a complete description of the \emph{status quo} policy can be found in~\Cref{appendix:statusquo}.

The results are shown in~\Cref{table:statusquo}. We consider two further variations of the \emph{status quo} policy. First, under $\delta$-tiebreaking, we sort patients of the same tier based on~\eqref{eq:delta}, with higher priority given to ones with higher $\delta$; as is evident from~\Cref{table:statusquo}, ties are quite common. The second variant, which can be used independently from the first one, excludes all transplants in which the selected pair incurs a negative benefit per~\eqref{eq:delta}; we found that this was quite common under the \emph{status quo} policy, and excluding such pairs turns out to have a considerable impact. For a fair comparison, the maximum distance in the myopic policy is taken to be $2000$ nautical miles; in the next two subsections, we will use the more conservative bound of $1000$ nautical miles.

\begin{table}[h!]
\centering
\footnotesize
\caption{Comparison with the \emph{status quo} policy in terms of life years gained. We report the average and standard deviation over $10$ independent runs for each month-policy combination.}
\begin{tabular}{@{}lccccc@{}}
\toprule
 & \multicolumn{5}{c}{\textbf{Policy}} \\
\cmidrule(l){2-6}
\textbf{Month} & Myopic & \emph{Status quo} & + $\delta$-exclusion & + $\delta$-tiebreaking & + $\delta$-exclusion + $\delta$-tiebreaking \\
\midrule
January 2019 & $1317 \pm 33$ & $932 \pm 47$ & $1151 \pm 33$ & $952 \pm 54$ & $1236 \pm 40$ \\
February 2019 & $1638 \pm 30$ & $1081 \pm 64$ & $1289 \pm 40$ & $1257 \pm 55$ & $1523 \pm 27$ \\
March 2019 & $1644 \pm 46$ & $1039 \pm 47$ & $1380 \pm 50$ & $1149 \pm 63$ & $1588 \pm 42$ \\
\bottomrule
\end{tabular}
\label{table:statusquo}
\end{table}

\begin{tcolorbox}[mytakeaway]
\textbf{Takeaway:} The \emph{status quo} policy fails to adequately account for pretransplant and post-transplant outcomes based on our simulations.
\end{tcolorbox}

In its defense, the status quo policy intends to account more for waitlist mortality, which partly explains its poor performance in terms of life years gained.

\subsection{A non-myopic policy: incorporating potentials}
\label{sec:potentials}

Although the myopic policy performs better than the \emph{status quo} in our simulations, the obvious concern is that it does not take into account the long-term composition of the pool. We address this by using \emph{potentials}, building on an existing approach in the context of kidney exchange~\citep{Dickerson12:Dynamic,Dickerson15:FutureMatch}. As a proof of concept, we only consider the blood type of the patient to construct the potential. We then select the patient that maximizes
\begin{equation}
\deltapot^{(t)} = \graftsurv(\donor^{(t)}, \pat ) - \waitsurv( \pat ) + \vec{\theta}^\top \vxpot,     
\end{equation}
where $\vec{\theta} = ( \theta_{O}, \theta_{A}, \theta_{B}, \theta_{AB})$ and $\vxpot$ contains the indicators for the corresponding blood types of the patient. What remains is to estimate $\vec{\theta}$. To do so, we train using the trajectories obtained from January $2019$ until March $2019$. We rely on a longer $3$-month time frame for training to make sure there is no overfitting; it is likely that going beyond $3$ months could improve performance even further, but we did not attempt to do so on account of computational restrictions. We use \texttt{SMAC} to estimate $\vec{\theta}$, a state-of-the-art package for algorithm configuration~\citep{Hutter11:Sequential}. We train for $50$ iterations of~\texttt{SMAC}. This experiment is more computationally intensive. Because of computational restrictions, we assume all offers get accepted, so, for a give month, the output is deterministic and there is no need to run multiple iterations of the simulator. The results are gathered in~\Cref{table:potential}. The potential-based policy delivers $31$ more years per month on average for the $9$ months tested, which is the equivalent of multiple additional transplants.


\begin{table}[h!]
\centering
\small
\caption{Myopic policy vs potential-based policy in terms of life years gained.}
\begin{tabular}{lcc}
\toprule
& \textbf{Myopic policy} & \textbf{Potential-based policy} \\
\midrule
April $2019$ & $1926$ & $1952$ \\
May $2019$ & $2174$ & $2219$ \\
June $2019$ & $1865$ & $1903$ \\
July $2019$ & $2063$ & $2104$ \\
August $2019$ & $1934$ & $1995$ \\
September $2019$ & $1927$ & $1967$ \\
October $2019$ & $2149$ & $2175$ \\
November $2019$ & $1882$ & $1867$ \\
December $2019$ & $1663$ & $1681$ \\
\bottomrule
\end{tabular}
\label{table:potential}
\end{table}

\begin{tcolorbox}[mytakeaway]
\textbf{Takeaway:} Incorporating potentials based on blood types significantly improves performance based on our simulations.
\end{tcolorbox}

\subsection{Batching multiple donors}
\label{sec:batch}

A key premise so far has been that the dispatch proceeds immediately after a donor has arrived. This is essential for donors who succumbed to \emph{circulatory death}, in which case the allocation has to proceed without any further delays. But the situation is different for \emph{brain dead} donors, who can be maintained in stable condition for a short interval. The main question we explore in this subsection pertains to the effect of allocating DBD donors in small batches; intuitively, this slightly alleviates the online nature of the problem. In particular, for a parameter $B \in \mathbb{N}$, we only proceed to allocation either when we have reached $B$ (DBD) donors, or when we have exceeded more than $48$ hours with respect to the first donor in the batch. (This time window was suggested by a professor of cardiothoracic surgery, who is our collaborator in the project.) Our main assumption is that those donors will remain stable for that duration.\footnote{In practice, we would recommend using more dynamic dispatch mechanisms: one can monitor the state of the organ and immediately proceed to dispatch as soon as there is noticeable deterioration in its key metrics.} Thereupon, we allocate based on the optimal myopic solution; this boils down to a maximum (weighted) bipartite matching problem---between the patients in the waitlist and the DBD donors in the batch---which can be solved optimally using, for example, the Hungarian algorithm. As before, in this subsection we assume that all offers proceed to transplant. The results are shown in~\Cref{table:batch}.

\begin{table}[h!]
\centering
\caption{The effect of batching DBD donors in terms of life years gained.}
\begin{tabular}{@{}lccccc@{}}
\toprule
 & \multicolumn{5}{c}{\textbf{Size of batch $B$}} \\
\cmidrule(l){2-6}
\textbf{Month} & 1 & 2 & 3 & 4 & 5 \\
\midrule
\multirow{2}{*}{January 2019} 
    & $1544$ & $1551$ & $1541$ & $1581$ & $1583$ \\
    & ---  & $7$ & $-3$ & $37$ & $39$ \\
\midrule
\multirow{2}{*}{February 2019} 
    & $1890$ & $1874$ & $1887$ & $1909$ & $1910$ \\
    & ---  & $-16$ & $-3$ & $19$ & $20$ \\
\midrule
\multirow{2}{*}{March 2019} 
    & $1848$ & $1868$ & $1881$ & $1903$ & $1905$ \\
    & ---  & $20$ & $33$ & $55$ & $57$ \\
\bottomrule
\end{tabular}
\label{table:batch}
\end{table}

\begin{tcolorbox}[mytakeaway]
\textbf{Takeaway:} Batching multiple DBD donors can significantly improve overall performance even for modest batch sizes (up to $5$) based on our simulations.
\end{tcolorbox}

The results reported in~\Cref{table:batch} are obtained without the use of potentials (per~\Cref{sec:potentials}). Combining potentials with batching is a natural direction, but training the potentials in that setting is computationally intensive, which is why we did not perform that experiment.
\section{Limitations}
\label{sec:limitations}

In this section, we highlight several limitations of our work. First, all our results are based on data provided by UNOS. It has to be cautioned that the UNOS registry is susceptible to errors in data entry and especially missing data. While we took some measures to detect and amend some incongruencies, we recognize that there are inherent limitations that we cannot address. More important, the predictors we develop in~\Cref{sec:simulator}---upon which our policy optimization relies---are highly imperfect. Problems such as predicting graft survival after a heart transplant are notoriously challenging; there may be confounders that were not present in the data and were thus not taken into account in the analysis, such as recipient psychosocial factors and gut microbiome composition. Going forward, it will be important to somehow account for the imperfections in the prediction models during the policy optimization phase. One advantage of our framework is that policy optimization is to a large extent orthogonal from the ML predictors; as a result, future developments in constructing more accurate models can be coupled directly with our optimization framework.

Regarding the optimization phase, we do not believe there is a definite objective for allocating heart transplants. We argue that the expected number of life years gained by transplantation is a natural objective, which is why we focus on that objective in this paper. But there are other important objectives one could consider, and our framework can be adapted accordingly.\footnote{An article by the~\citet{optn_ethical} discusses at greater length such complex ethical considerations.} Our goal in this paper is not to make policy recommendations. Rather, we are developing the framework and tools that will enable stakeholders to make informed decisions concerning different tradeoffs.

Finally, one simplifying assumption we made in the development of our simulator is that certain patients will remain static for the duration of the simulation. Addressing this assumption will necessitate developing accurate models for predicting patient progression in the waitlist, which is an interesting direction for future work.
\section{Conclusions and future research}

Notwithstanding those limitations, we made an important step toward improving the \emph{status quo} policy for heart transplant allocation, which would have considerable societal impact. Future research should focus primarily on two distinct threads. First, the policies obtained from our framework are only as good as the underlying predictors. There is certainly room for improving those models and developing a more reliable simulator. The second promising direction concerns the policy optimization step itself. We saw that incorporating potentials based on patient blood types led to a considerable improvement in the performance of the policy \emph{vis-\`a-vis} the myopic one. Going forward, it will be important to take other covariates into account when setting up the potentials. The main bottleneck here is computational: training the potentials is particularly expensive, so new ideas are required in order to tackle higher dimensional formulations. Finally, the main objective we consider in this paper quantifies the number of (estimated) years gained through the transplants. It is likely that this objective is not always aligned with fairness with respect to different demographic groups, for example, based on ethnicity. Another common risk group in transplantation is highly sensitized patients, for whom it is difficult to find a matching organ. One natural fairness notion is the max-min fairness objective, which measures performance on the group that is worst off. Extending our framework to address fairness considerations is another important direction for the future.

\section*{Acknowledgments}

Tuomas Sandholm is supported by the Vannevar Bush Faculty Fellowship ONR N00014-23-1-2876, National Science Foundation grants RI-2312342 and RI-1901403, ARO award W911NF2210266, and NIH award A240108S001. We are indebted to Carlos Martinez from UNOS for answering countless of our questions.

\bibliography{dairefs}

\appendix

\section{\emph{Status quo} heart transplant allocation policy}
\label{appendix:statusquo}

The current heart transplant allocation policy is based on a hierarchical prioritization system that considers three primary factors: medical urgency status, blood type compatibility, and geographic proximity. For completeness, this section spells out how the priority tiers are defined; we reiterate that we only consider adult donors and patients.

The allocation system assigns patients to 68 distinct priority tiers; 1 represents the highest priority, while 68 the lowest. Each tier is defined by a unique combination of three criteria:

\begin{itemize}
    \item Medical status, ranging from status 1 (most urgent) to status 6 (least urgent).
    \item Blood type compatibility:
    \begin{itemize}
        \item If the donor has blood type $O$, patients with blood type $O$ or $B$ have \emph{primary} blood compatibility. The rest of the patients have \emph{secondary} blood compatibility.
        \item If the donor has blood type $A$, patients with blood type $A$ or $AB$ have primary blood compatibility. The rest of the patients are not compatible.
        \item If the donor has blood type $B$, patients with blood type $B$ or $AB$ have primary blood compatibility. The rest of the patients are not compatible.
        \item If the donor has blood type $AB$, patients with blood type $AB$ have primary blood compatibility. The rest of the patients are not compatible.
    \end{itemize}
    \item Distance: geographic distance in nautical miles (nm) between donor and recipient locations, giving rise to $6$ different zones.
\end{itemize}

Table~\ref{tab:priority_tiers} presents the complete 68-tier priority system. Within each tier, patients are further ranked by decreasing waiting time or other secondary criteria.

\begin{table}[htbp]
\footnotesize
\centering
\caption{\emph{Status quo} priority tiers.}
\label{tab:priority_tiers}
\begin{tabular}{clll@{\hspace{1em}}|@{\hspace{1em}}clll}
\toprule
\textbf{Tier} & \textbf{Status} & \textbf{Blood match} & \textbf{Distance (nm)} & \textbf{Tier} & \textbf{Status} & \textbf{Blood match} & \textbf{Distance (nm)} \\
\midrule
1 & 1 & Primary & $\leq 500$ & 35 & 2 & Primary & $\leq 2500$ \\
2 & 1 & Secondary & $\leq 500$ & 36 & 2 & Secondary & $\leq 2500$ \\
3 & 2 & Primary & $\leq 500$ & 37 & 3 & Primary & $\leq 2500$ \\
4 & 2 & Secondary & $\leq 500$ & 38 & 3 & Secondary & $\leq 2500$ \\
5 & 3 & Primary & $\leq 250$ & 39 & 4 & Primary & $\leq 1000$ \\
6 & 3 & Secondary & $\leq 250$ & 40 & 4 & Secondary & $\leq 1000$ \\
7 & 1 & Primary & $\leq 1000$ & 41 & 5 & Primary & $\leq 1000$ \\
8 & 1 & Secondary & $\leq 1000$ & 42 & 5 & Secondary & $\leq 1000$ \\
9 & 2 & Primary & $\leq 1000$ & 43 & 6 & Primary & $\leq 1000$ \\
10 & 2 & Secondary & $\leq 1000$ & 44 & 6 & Secondary & $\leq 1000$ \\
11 & 4 & Primary & $\leq 250$ & 45 & 1 & Primary & Any \\
12 & 4 & Secondary & $\leq 250$ & 46 & 1 & Secondary & Any \\
13 & 3 & Primary & $\leq 500$ & 47 & 2 & Primary & Any \\
14 & 3 & Secondary & $\leq 500$ & 48 & 2 & Secondary & Any \\
15 & 5 & Primary & $\leq 250$ & 49 & 3 & Primary & Any \\
16 & 5 & Secondary & $\leq 250$ & 50 & 3 & Secondary & Any \\
17 & 3 & Primary & $\leq 1000$ & 51 & 4 & Primary & $\leq 1500$ \\
18 & 3 & Secondary & $\leq 1000$ & 52 & 4 & Secondary & $\leq 1500$ \\
19 & 6 & Primary & $\leq 250$ & 53 & 5 & Primary & $\leq 1500$ \\
20 & 6 & Secondary & $\leq 250$ & 54 & 5 & Secondary & $\leq 1500$ \\
21 & 1 & Primary & $\leq 1500$ & 55 & 6 & Primary & $\leq 1500$ \\
22 & 1 & Secondary & $\leq 1500$ & 56 & 6 & Secondary & $\leq 1500$ \\
23 & 2 & Primary & $\leq 1500$ & 57 & 4 & Primary & $\leq 2500$ \\
24 & 2 & Secondary & $\leq 1500$ & 58 & 4 & Secondary & $\leq 2500$ \\
25 & 3 & Primary & $\leq 1500$ & 59 & 5 & Primary & $\leq 2500$ \\
26 & 3 & Secondary & $\leq 1500$ & 60 & 5 & Secondary & $\leq 2500$ \\
27 & 4 & Primary & $\leq 500$ & 61 & 6 & Primary & $\leq 2500$ \\
28 & 4 & Secondary & $\leq 500$ & 62 & 6 & Secondary & $\leq 2500$ \\
29 & 5 & Primary & $\leq 500$ & 63 & 4 & Primary & Any \\
30 & 5 & Secondary & $\leq 500$ & 64 & 4 & Secondary & Any \\
31 & 6 & Primary & $\leq 500$ & 65 & 5 & Primary & Any \\
32 & 6 & Secondary & $\leq 500$ & 66 & 5 & Secondary & Any \\
33 & 1 & Primary & $\leq 2500$ & 67 & 6 & Primary & Any \\
34 & 1 & Secondary & $\leq 2500$ & 68 & 6 & Secondary & Any \\
\bottomrule
\end{tabular}
\end{table}

\end{document}